\documentclass[11pt]{article}

\usepackage[preprint]{acl}

\usepackage{times}
\usepackage{latexsym}

\usepackage[T1]{fontenc}
\usepackage[utf8]{inputenc}

\usepackage{microtype}

\usepackage{inconsolata}

\usepackage{kotex}
\usepackage{xspace}
\usepackage{verbatim}
\usepackage{fancyvrb}
\usepackage{fvextra}
\usepackage{multirow}
\usepackage{booktabs} 
\usepackage{arydshln}
\usepackage{pifont}
\usepackage{algpseudocode}
\usepackage{xcolor}
\usepackage{amsmath}
\usepackage{amssymb}
\usepackage{wasysym}
\usepackage{makecell}
\usepackage{colortbl}
\usepackage{tcolorbox}
\tcbuselibrary{listings, breakable, skins}
\usepackage{fancyvrb}
\usepackage{graphicx}
\usepackage{tabularx}
\usepackage{subcaption} 
\usepackage{array}
\usepackage{placeins}

\newcolumntype{Y}[1]{>{\raggedright\arraybackslash}m{#1}}
\newcolumntype{C}[1]{>{\centering\arraybackslash}m{#1}}

\newtcolorbox{MyPromptBox}{
  breakable,
  colback=gray!5,
  colframe=black!50,
  boxrule=0.5pt,
  arc=2pt,
  left=6pt, right=6pt, top=6pt, bottom=6pt,
  fontupper=\small\ttfamily
}

\newcommand{\mymethod}[0]{\textsc{AMuFC}\xspace}

\title{Is a Picture Worth a Thousand Words?\\Adaptive Multimodal Fact-Checking with Visual Evidence Necessity}

\author{
Jaeyoon Jung$^{\spadesuit\diamondsuit}$~~~~~~Yejun Yoon$^{\heartsuit}$~~~~~~Kunwoo Park$^{\spadesuit\heartsuit}$\\ 
$^{\spadesuit}$School of AI Convergence, Soongsil University\\
$^{\diamondsuit}$MAUM AI Inc.\\
$^{\heartsuit}$Department of Intelligent Semiconductors, Soongsil University\\
\texttt{\small\{jaeyoonskr, yejun0382\}@soongsil.ac.kr, kunwoo.park@ssu.ac.kr}
}

\begin{document}
\maketitle
\begin{abstract}
  Automated fact-checking is a crucial task that supports a responsible information ecosystem. While recent research has progressed from text-only to multimodal fact-checking, a prevailing assumption is that incorporating visual evidence universally improves verification accuracy. In this work, we challenge this assumption and show that the indiscriminate use of visual evidence can reduce accuracy. Building on this finding, we propose \mymethod, a modular fact-checking framework that employs two collaborative vision-language models with distinct roles to enable the adaptive use of visual evidence. Experimental results on three datasets, including WebFC, introduced in this study, demonstrate the effectiveness of adaptive visual evidence use in fact-checking. 
\end{abstract}

\section{Introduction}

Fact-checking---the process of determining the veracity of claims---is a cornerstone of responsible journalism and an essential mechanism for mitigating the harms of misinformation in broader web environments~\cite{hassan2015detecting,adair2017progress}. Building on early work focused on text-only verification, recent research has increasingly shifted toward multimodal fact-checking, which addresses scenarios in which evidence is conveyed through multiple modalities, primarily text and images~\cite{hameleers2020picture, alam-etal-2022-survey, biamby-etal-2022-twitter}. Various approaches have been proposed to integrate these modalities for fact verification. For example, multimodal fusion-based classifiers combine retrieved image--text evidence~\cite{mocheg}, while more recent work leverages reinforcement learning to summarize multimodal inputs~\cite{metasumperceiver}. A common finding across these studies is that leveraging visual evidence can improve fact-checking accuracy~\cite{misinformation,mmfc-vlm}, echoing the adage that \emph{a picture is worth a thousand words}. 

\begin{figure}[t]
    \centering
    \includegraphics[width=.99\linewidth]{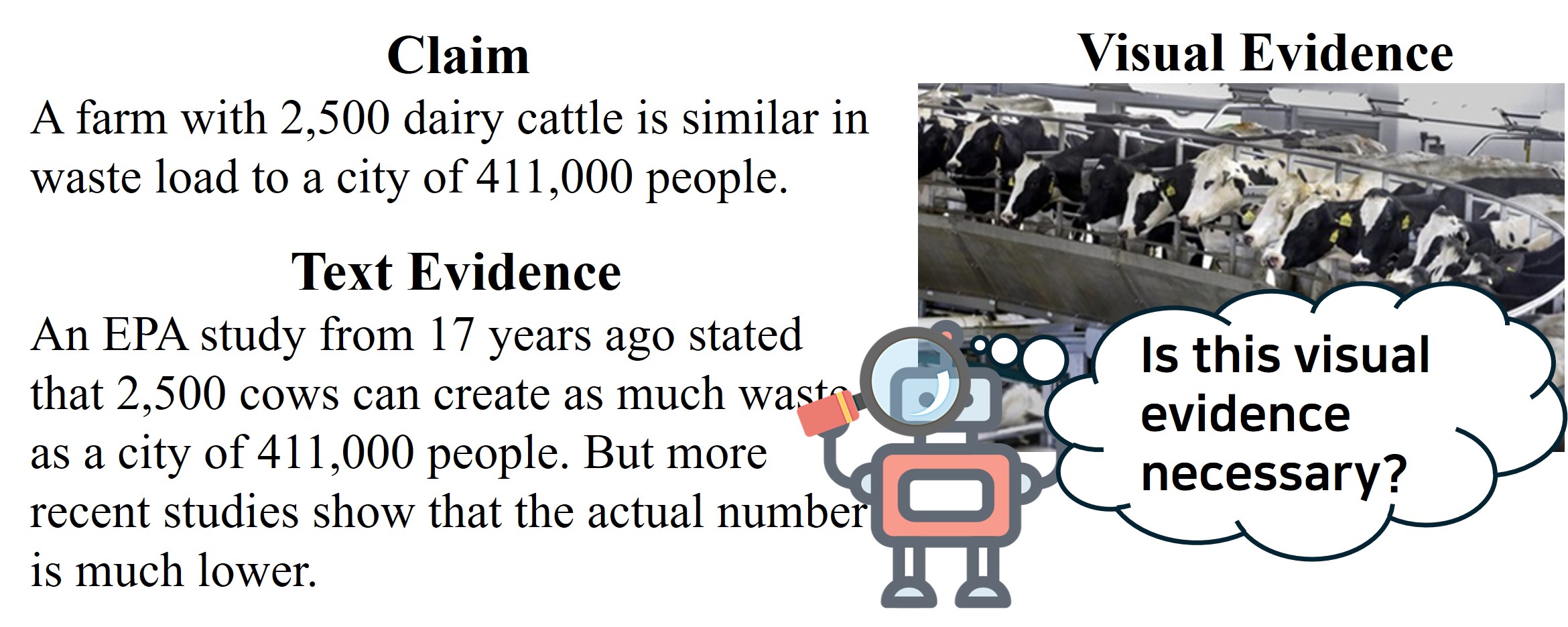}
    \caption{An example illustrating that visual evidence is not always necessary for multimodal fact-checking, motivating the research direction of this study.}
    \label{fig:example_of_fig0}
\end{figure}

\begin{table*}[t]
\centering
\resizebox{0.99\textwidth}{!}{%
\begin{tabular}{lccccccc}
\toprule
\textbf{Dataset} & \makecell{\textbf{Primary}\\\textbf{Inputs}} & \textbf{Evidence} & \textbf{Task} &
\makecell{\textbf{Label}\\\textbf{Source}} &
\makecell{\textbf{Claim}\\\textbf{Source}} &
\makecell{\textbf{External}\\\textbf{Knowledge}\\\textbf{Source}} &
\textbf{Access} \\
\midrule
Fakeddit~\cite{nakamura-etal-2020-fakeddit}  & Img/Txt & -  & Mixed & Synthetic & Reddit & \ding{55} & \ding{51} \\
NeuralNews~\cite{tan-etal-2020-detecting} & Img/Txt & - & OOC & Mixed & Grover/GoodNews & \ding{55} & \ding{51}\\
NewsCLIPings~\cite{luo-etal-2021-newsclippings} & Img/Txt & - & OOC & Mixed & CLIP/VisualNews & \ding{55} & \ding{51} \\
Factify~\cite{suryavardan2023factify} & Img/Txt & Img/Txt & OOC & Mixed & Twitter & \ding{55} & \ding{51}\\
VERITE~\cite{papadopoulos2024verite} & Img/Txt & - & Mixed & Mixed & Snopes/Reuters & \ding{55} & \ding{51}\\
Fauxtography~\cite{zlatkova-etal-2019-fact} & Img/Txt & Meta & FV & Expert & Snopes/Reuters & \ding{55} & \ding{51} \\
MMM~\cite{gupta2022mmm} & Img/Txt & Txt & FV & Mixed & FC webs & \ding{55} & \ding{55} \\
FactDrill~\cite{singhal2022factdrill} & Txt & Vid/Aud/Img/Txt/Meta & FV & Expert & FC webs & \ding{55} & \ding{55} \\
ChartFC~\cite{akhtar-etal-2023-reading} & Txt & Img & FV & Crowd & TabFact & \ding{55} & \ding{51}\\
ChartCheck~\cite{chartcheck} & Txt & Img/Txt & FV & Crowd & Wikimedia & \ding{55} & \ding{51} \\
MOCHEG~\cite{mocheg} & Txt & Img/Txt & FV & Expert & FC webs & \ding{51} & \ding{51} \\
FIN-FACT~\cite{rangapur2025fin} & Txt & Img/Txt & FV & Expert & FC webs & \ding{55} & \ding{51} \\
AVerImaTeC~\cite{cao2025averimatec} & Img/Txt & Img/Txt & FV & Expert & FC webs & \ding{51} & \ding{51}\\
VeriTaS~\cite{rothermel-etal-2026-veritas} & Vid/Img/Txt & - & FV & Mixed & FC webs & \ding{55} & \ding{51}\\
\bottomrule
\end{tabular}
}
\caption{Datasets for multimodal fact-checking (OOC: out-of-context misinformation; FV: fact verification).}
\label{tab:dataset_table}
\end{table*}

In this study, we challenge a prevailing assumption in prior work---the presumed necessity of visual evidence---as illustrated in Figure~\ref{fig:example_of_fig0}. Using MOCHEG~\cite{mocheg} as our primary testbed, we analyze the effects of visual evidence on verification accuracy. While prior work has shown that some claims cannot be reliably verified without visual evidence~\cite{mmfc-vlm}, we show that a na\"ive strategy of indiscriminately incorporating visual evidence can lead models to underperform text-only baselines. We further show that verification performance improves when the evidence modality is selected adaptively, as demonstrated in an oracle setting in which visual evidence is incorporated only when it is beneficial for verification. 

Motivated by these findings, we propose \mymethod, a modular framework that adaptively incorporates visual evidence through the collaboration of two vision-language models (VLMs) with distinct roles. To examine whether the benefits of adaptive visual evidence use persist in practical settings, we adopt a simple Analyzer--Verifier design. An \emph{Analyzer} determines whether visual evidence is necessary for verification, while a \emph{Verifier} predicts claim veracity based on the Analyzer's natural-language assessment. Experimental results on MOCHEG demonstrate that \mymethod outperforms existing methods. We further conduct test-only evaluations on FIN-FACT~\cite{rangapur2025fin}, a financial-domain dataset, and WebFC, a newly constructed dataset of recent claims paired with web-retrieved evidence, and observe that the performance gains from incorporating the Analyzer's assessment persist. These results suggest that the adaptive use of visual evidence, as operationalized by \mymethod, can be effective across diverse fact-checking scenarios, including cross-domain verification and verification of recent claims using noisy web evidence.

The key contributions of this study are summarized below.
\begin{itemize}
\item We show that visual evidence is not always beneficial for multimodal fact-checking, challenging a prevailing assumption in prior research that incorporating visual evidence inherently improves verification performance.
\item We propose \mymethod, a modular framework that adaptively uses visual evidence. Despite its simplicity, \mymethod demonstrates the effectiveness of adaptive visual evidence use across diverse fact-checking scenarios.
\end{itemize}

\section{Related Work}
\subsection{Multimodal Fact-checking}

Previous research on multimodal fact-checking, or multimodal misinformation detection, can be broadly categorized into three tasks based on their verdict prediction settings: \emph{manipulation classification}, \emph{out-of-context detection}, and \emph{fact verification}~\cite{akhtar-etal-2023-multimodal}. Manipulation classification focuses on detecting manipulated misinformation within a single modality~\cite{gupta2013faking,boididou2014challenges,huh2018fighting,heller2018ps,shao2023detecting}. Out-of-context detection aims to determine whether a given image-text pair conveys consistent contextual information~\cite{nakamura-etal-2020-fakeddit,tan-etal-2020-detecting,luo-etal-2021-newsclippings,suryavardan2023factify,papadopoulos2024verite}. Fact verification, which is the focus of this paper, seeks to predict the truthfulness of a claim, typically in textual form~\cite{singhal2022factdrill,akhtar-etal-2023-reading,chartcheck,mocheg}. Table~\ref{tab:dataset_table} summarizes widely used multimodal fact-checking datasets. 

Following early attempts at zero-shot fact verification~\cite{lee-etal-2020-language}, most subsequent work has adopted retrieval-augmented generation (RAG) frameworks~\cite{lewis2020retrieval}, which retrieve both visual and textual evidence and leverage the retrieved content to support verification.
Because evidence quality is critical for RAG-based verification, several studies have focused on mitigating noisy or irrelevant context. For example, MetaSumPerceiver was proposed to summarize multimodal document evidence~\cite{metasumperceiver}. Other work introduced re-ranking methods based on predicted token probabilities~\cite{misinformation}, while additional approaches proposed tool-based RAG pipelines for fact verification. Most recently, \citet{Pang_Li_Zhang_Wang_Zhang_2025} proposed a hypergraph transformer-based fact-checking framework to capture higher-order relationships between claims and evidence. Our study explicitly examines the necessity of visual evidence for fact verification.

\subsection{Adaptive Retrieval}

We review previous research on adaptive retrieval for fact verification and, more broadly, for RAG. Although RAG improves performance on knowledge-intensive tasks~\cite{lewis2020retrieval}, inaccurate or irrelevant retrieval can introduce noisy context that degrades performance~\cite{yoran2024making,shi2023large}. Adaptive retrieval strategies mitigate this issue by retrieving on demand or filtering out unhelpful context~\cite{tang-etal-2025-mba,parekh-etal-2025-dynamic}. For example, Self-RAG~\cite{asai2023self} enables dynamic retrieval via reflection tokens, while other approaches select RAG strategies based on query complexity~\cite{adaptive-rag} or use trained relevance estimators to assess context utility~\cite{kim-lee-2024-rag}. More recent work leverages hidden-state representations to decide whether retrieval is necessary~\cite{baek-etal-2025-probing} or dynamically incorporates retrieved passages based on their quality~\cite{in-etal-2025-diversify}. 

In fact verification, agent-based iterative frameworks integrate retrieval and verification across multiple rounds~\cite{xie-etal-2025-fire}, while PASS-FC augments claims with temporal and entity grounding for progressive search~\cite{zhuang2025pass}. Related work also explores dynamically adjusting search depth to extract multimodal evidence~\cite{braun2025defame}. In contrast to prior work that primarily focuses on textual relevance or filters out irrelevant evidence before verification, this study focuses on the necessity of visual evidence and introduces a modular framework that enables the adaptive use of visual information.

\section{Task and Dataset}
\label{sec:dataset}

Given a textual claim $c$ and an external knowledge source $K$, the task of \emph{multimodal fact verification} aims to determine the veracity of $c$ by retrieving and reasoning over both textual and visual evidence. Specifically, a fact-checking system retrieves a set of textual evidence $E_t\subset K$ and visual evidence $E_v\subset K$, and then predicts a verdict label $y \in \{$\emph{supported}, \emph{refuted}, \emph{not enough information (NEI)}$\}$ based on $(c,E_t,E_v)$.

Among existing multimodal fact-checking datasets (Table~\ref{tab:dataset_table}), we select MOCHEG~\cite{mocheg} as the primary testbed for our study based on three criteria aligned with our research objectives. First, the dataset must explicitly target \emph{fact verification}, thereby excluding datasets designed for related but distinct tasks, such as out-of-context detection in Factify~\cite{suryavardan2023factify}. Second, it must take a textual claim $c$ as input, excluding datasets such as AVerImaTeC~\cite{cao2025averimatec} that assume text--image input pairs. Third, it must provide a knowledge source $K$ from which textual and visual evidence can be retrieved. MOCHEG~\cite{mocheg} satisfies all three criteria and is therefore selected as our primary testbed. To assess generalizability beyond a single dataset, we use two additional datasets exclusively for testing: FIN-FACT and WebFC. These datasets represent diverse fact-checking scenarios, including generalization to claims from other domains and to recent claims, without in-domain training.

The datasets used in this study are described below. 
\begin{itemize}
    \item \textbf{MOCHEG} comprises claims with corresponding verdicts, supported by evidence curated by professional fact-checkers. The claims are collected from two fact-checking websites: PolitiFact\footnote{\url{https://www.politifact.com/}} and Snopes\footnote{\url{https://www.snopes.com/}}. The dataset includes 11,669 claims for training, 1,490 for validation, and 2,442 for testing. The knowledge source $K$ consists of 3,070,563 sentences and 137,621 images. Each claim is paired with gold textual evidence and, when available, gold visual evidence.
    \item \textbf{FIN-FACT}~\cite{rangapur2025fin} is a financial fact-checking dataset containing claims and verdicts produced by human experts. We applied a sanitization process to the original dataset by removing claims without publicly accessible image evidence and refining the remaining claims, yielding 2,581 claims with associated multimodal evidence. As the dataset does not provide an external knowledge source, we treat the entire set of textual evidence and images as $K$.
    \item \textbf{WebFC} is a newly constructed corpus developed in this study, consisting of 621 recently published claims and their corresponding verdicts from PolitiFact, spanning January 2024 to September 2025. Treating web documents as the knowledge source $K$, we retrieved both textual and visual evidence using the Google Custom Search API. To prevent the retrieval of documents published after the corresponding fact-checking articles, which would create an unrealistic verification setting, we restricted retrieval to sources published before the fact-checking articles, following recent practice~\cite{cao2025averimatec}. We also excluded evidence from fact-checking organizations to reduce the risk of directly retrieving previously published verdicts. Accordingly, WebFC better reflects a temporally realistic fact-checking scenario in which recent claims are verified using web-retrieved evidence, while also reducing the risk of temporal leakage from evidence published after the corresponding fact-checking decisions.
\end{itemize}

Details of dataset preprocessing and construction are provided in Appendix~\ref{section:dataset_detail}. 

\section{Is Visual Evidence Always Needed?}
\label{section:4}

We challenge a common assumption in prior research on multimodal fact-checking---that visual evidence is necessary for effective verification. Using MOCHEG, we examine how incorporating visual evidence affects fact-checking performance and then conduct qualitative analyses to identify when visual evidence contributes to verification and when it is unnecessary.

\subsection{Effect of Visual Evidence on Performance}
\label{subsection:4.1}

We compared the verification accuracy of VLMs across four input configurations: (1) textual evidence only, (2) textual evidence with \emph{gold} image evidence, (3) textual evidence with retrieved image evidence, and (4) an oracle setting in which an input configuration yielding the correct verdict was selected for each instance, whenever one existed. 

In all configurations, we assumed that the \emph{gold} textual evidence was provided, as our focus was to assess the impact of visual evidence. Following prior work~\cite{mmfc-vlm}, configuration (2) used the first image as evidence when multiple images were available. In configuration (3), we used the fine-tuned CLIP model provided with the MOCHEG baseline to retrieve the top-1 image from $K$ as visual evidence. In configuration (4), the input combination that yielded the correct verdict---when such a configuration existed---was always selected, thereby representing an upper bound on achievable performance under perfect modality selection. This controlled comparison enabled us to isolate the impact of visual evidence on fact verification performance.

Table~\ref{tab:claim_verification_performance} presents the evaluation results. The first column lists the target VLMs, the second specifies the input configuration, and the third and fourth report verification performance on the MOCHEG test split, measured by accuracy (ACC) and macro F1. We evaluated two open-weight and two API-based VLMs. The open-weight models are Qwen2-VL-7B~\cite{wang2024qwen2} and Llama-3.2-11B-Vision~\cite{grattafiori2024llama}, denoted throughout the paper as Qwen2-VL and Llama-3.2-V, respectively, while the API-based models are GPT-4o~\cite{hurst2024gpt} and Gemini-2.5-Pro~\cite{comanici2025gemini}. Model checkpoints are provided in Appendix~\ref{section:configurations}.

\begin{table}[t]
\centering
\resizebox{0.48\textwidth}{!}{%
\begin{tabular}{clcc}
\toprule
\textbf{Model} & \textbf{Configuration} & \textbf{ACC} & \textbf{F1} \\
\midrule

\multirow{4}{*}{Qwen2-VL}
& (1) Text & 0.509 & 0.419 \\
& (2) $+$ Image (G) & 0.506 & 0.410 \\
& (3) $+$ Image (R) & 0.490 & 0.395 \\
& (4) Oracle & \textbf{0.547} & \textbf{0.451} \\
\midrule

\multirow{4}{*}{Llama-3.2-V}
& (1) Text & 0.496 & 0.495 \\
& (2) $+$ Image (G) & 0.462 & 0.445 \\
& (3) $+$ Image (R) & 0.412 & 0.393 \\
& (4) Oracle & \textbf{0.590} & \textbf{0.582} \\
\midrule

\multirow{4}{*}{GPT-4o}
& (1) Text & 0.642(.001) & 0.612(.000) \\
& (2) $+$ Image (G) & 0.635(.003) & 0.596(.003) \\
& (3) $+$ Image (R) & 0.630(.001) & 0.597(.001) \\
& (4) Oracle & \textbf{0.680}(.002) & \textbf{0.649}(.001) \\
\midrule

\multirow{4}{*}{Gemini-2.5-Pro}
& (1) Text & 0.627(.001) & 0.599(.001) \\
& (2) $+$ Image (G) & 0.620(.001) & 0.570(.001) \\
& (3) $+$ Image (R) & 0.596(.001) & 0.557(.001) \\
& (4) Oracle & \textbf{0.692}(.001) & \textbf{0.660}(.001) \\
\bottomrule
\end{tabular}
}
\caption{Fact-verification performance across different evidence configuration strategies, indicating that visual evidence is not always necessary. Values in parentheses denote standard errors, which are omitted for open-weight models due to deterministic results (G: Gold, R: Retrieved).}
\label{tab:claim_verification_performance}
\end{table}

\renewcommand\tabularxcolumn[1]{m{#1}}
\begin{table*}[t]
\centering
\small
\setlength{\tabcolsep}{4pt}
\renewcommand{\arraystretch}{1.2}
\renewcommand\tabularxcolumn[1]{m{#1}} 
\begin{tabularx}{\textwidth}{c X X}
\toprule
\textbf{Type} & \makecell[c]{\textbf{Unnecessary}} & \makecell[c]{\textbf{Necessary}} \\
\midrule

\textbf{Claim} &
Marquette University threatened to rescind student's admission over pro-Trump TikTok video. &
In an episode of 'The Simpsons,' Mayor Quimby says he is canceling a trip to the Bahamas while he's in the Bahamas, because of an ongoing epidemic. \\
\midrule

\textbf{Visual Evidence} &
\makebox[\linewidth][c]{\includegraphics[width=\linewidth,height=3.5cm,keepaspectratio]{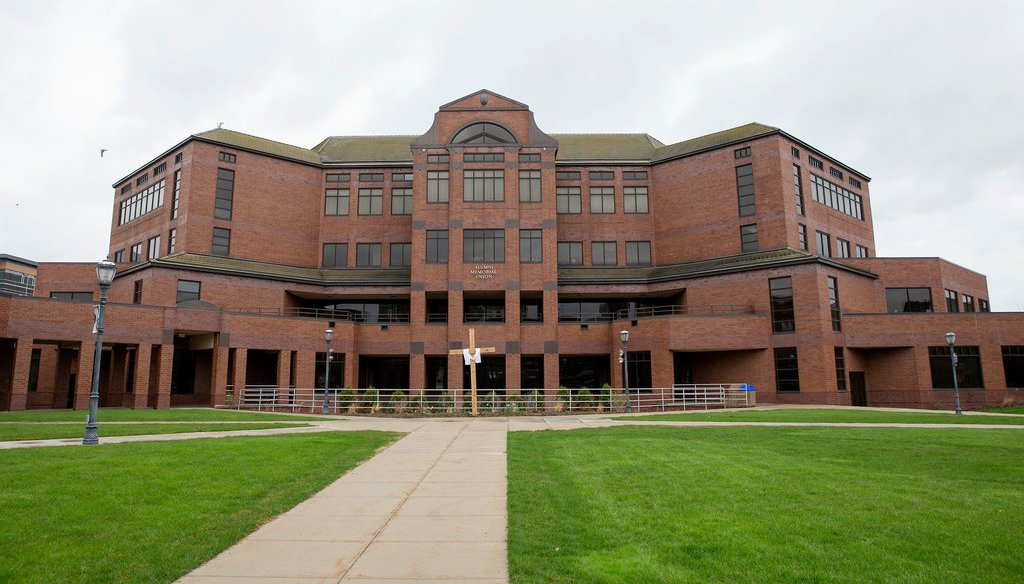}} &
\makebox[\linewidth][c]{\includegraphics[width=\linewidth,height=3.5cm,keepaspectratio]{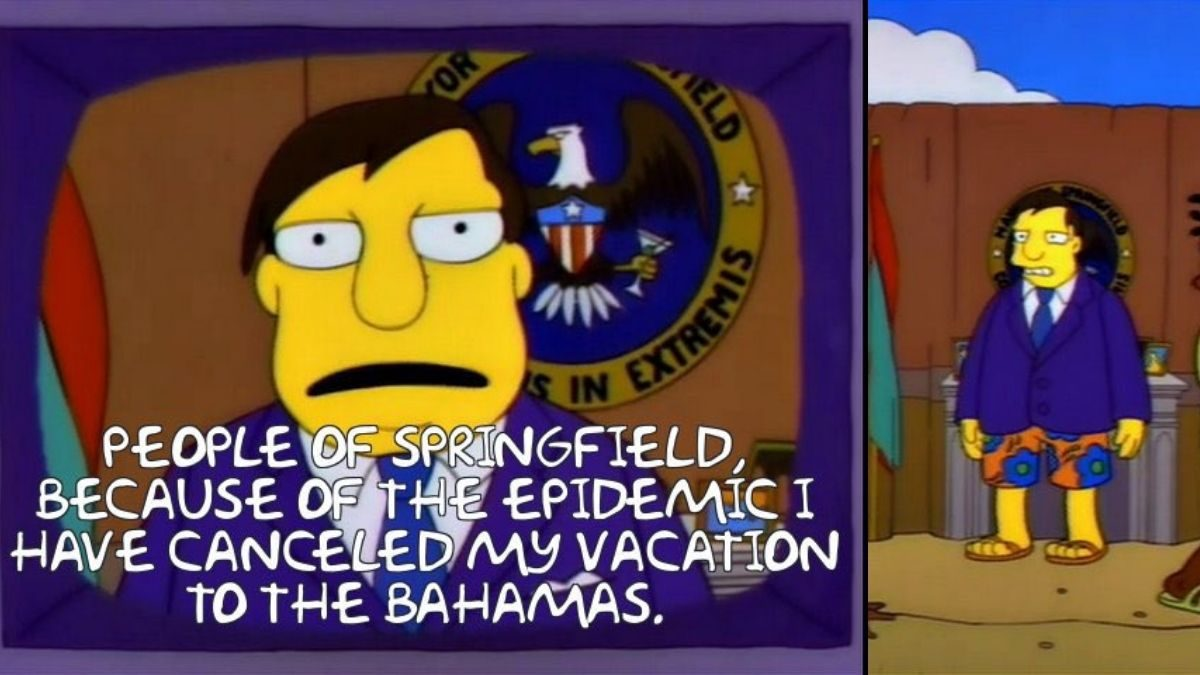}} \\
\midrule

\textbf{Textual Evidence} &
\hangindent=1em \noindent • The pro-Trump post was not at issue. Marquette and the people complaining to the university were examining comments on topics like sexuality and immigration in other social media posts. 

\hangindent=1em \noindent • The student clearly felt her status was in doubt, but she has stopped short of calling it a threat. The school says no threat was made.

\hangindent=1em \noindent • The “threat” element is tricky to pin down, since that claim stems from a private conversation between the student and Marquette admissions staffers. &

After initially saying he didn't do anything wrong, Steve Adler, the mayor of Austin, Texas, says he now realizes he 'set a bad example' by traveling to Cabo San Lucas, Mexico, for vacation last month. An Austin American-Statesman story revealed Wednesday that Adler attended an in-person wedding for his daughter in early November and then flew with others to Cabo for a weeklong vacation. At the same time, Adler was encouraging people to stay home to avoid contracting or spreading COVID-19. \\
\bottomrule
\end{tabularx}

\caption{Examples for visual evidence types according to their necessity for claim verification.}
\label{tab:visual_role}
\end{table*}

We derived three key observations regarding the effects of visual evidence on verification performance. First, when comparing the text-only setting (configuration 1) with the multimodal setting using gold images (configuration 2), incorporating visual evidence consistently degraded performance across all evaluated models. Second, in configuration (3), which uses retrieved image evidence, performance decreased relative to both configurations (1) and (2) across all models. The largest drop was observed for Llama-3.2-V, with decreases of 0.084 in accuracy and 0.102 in macro F1. This result suggests that imperfect retrieval can introduce noisy or misleading context. Third, the oracle setup (configuration 4) consistently outperformed all other settings through optimal modality selection. According to the Mann--Whitney U test, the improvement over the non-oracle settings was statistically significant (\textit{p}$<$0.01) for GPT-4o and Gemini-2.5-Pro.\footnote{The results for the open-weight models are deterministic and therefore were not included in the statistical test.} Taken together, these findings support our hypothesis that visual evidence is not always necessary for claim verification. Moreover, the substantial performance gains observed in the oracle setting motivate the adaptive use of visual evidence.

\subsection{When Does Visual Evidence Not Help?}
\label{subsection:4.2}

Building on the mixed effects of visual evidence on verification performance, we conducted a qualitative analysis to examine when visual evidence does not aid verification. We first defined \emph{claim categories} based on whether visual evidence changed GPT-4o's verification outcome on the validation set, as GPT-4o achieved the best overall performance among the four VLMs in Table~\ref{tab:claim_verification_performance}. Specifically, we labeled claims as \textbf{Visual-successful} (51 claims) when the model predicted the correct verdict only after visual evidence was added to the textual evidence, and as \textbf{Visual-unsuccessful} (61 claims) when the model predicted the correct verdict using textual evidence alone but changed to an incorrect prediction after visual evidence was incorporated. Since our goal is to understand the influence of visual evidence on verification, we excluded claims for which adding images did not affect the text-only model's prediction.

\begin{figure}[t]
    \centering
    \includegraphics[width=.98\linewidth]{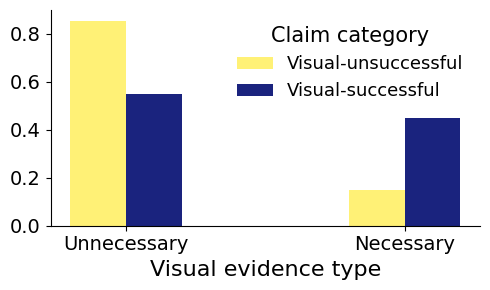}
    \caption{Distribution of visual evidence types across claim categories, illustrating the association between visual evidence necessity and verification outcomes.}
    \label{fig:example_of_fig4}
\end{figure}

\begin{figure*}[t]
    \centering
    \includegraphics[width=.98\linewidth]{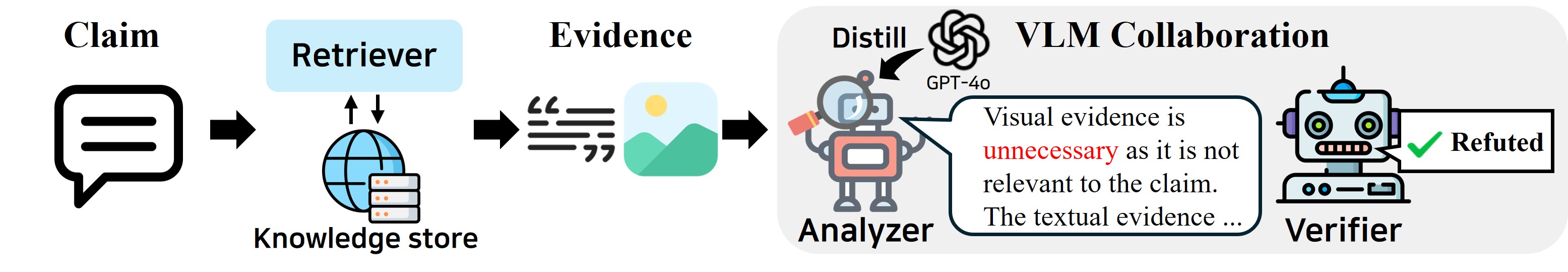}
    \caption{Overall pipeline of \mymethod. Given the retrieved evidence, the two VLMs---Analyzer and Verifier---are responsible for assessing the necessity of visual evidence and predicting the claim's veracity, respectively.}
    \label{fig:pipeline}
\end{figure*}

Next, we manually assessed whether visual evidence was necessary for claim verification. Through an iterative process, we developed an annotation scheme that distinguished two \emph{visual evidence types} according to their necessity for verification. We annotated visual evidence as \textbf{Unnecessary} (80 pieces) when it was redundant (i.e., it merely repeated information already conveyed by the text) or irrelevant (e.g., it omitted the key entity or event mentioned in the claim, or merely showed a portrait of a person without adding contextual value), and as \textbf{Necessary} (32 pieces) when it provided unique information required to evaluate the claim's truthfulness. Two annotators from Soongsil University applied this scheme to all 112 samples, whose claim categories had already been determined based on GPT-4o's predictions, following the annotation guidelines in Figure~\ref{fig:annotation_guideline}. This process achieved high agreement (Krippendorff's $\alpha = 0.809$), indicating substantial inter-annotator agreement. Examples are shown in Table~\ref{tab:visual_role}.

Figure~\ref{fig:example_of_fig4} shows how the manually annotated visual evidence types (Unnecessary or Necessary) varied across the automatically determined claim categories (Visual-successful or Visual-unsuccessful). We observed a significant association between claim categories and evidence types (\textit{p}$<$0.001, chi-square test), indicating that the effect of visual evidence on verification performance is closely related to whether the visual evidence is actually necessary for verification. In particular, redundant or irrelevant visual evidence may degrade verification performance rather than improve it. Consistent with the results in Table~\ref{tab:claim_verification_performance}, these findings motivate the adaptive use of visual evidence.

\section{Proposed Method: \mymethod}
\label{section:5}

We propose a modular fact-checking framework that adaptively incorporates visual evidence, termed \textbf{A}daptive \textbf{Mu}ltimodal \textbf{F}act-\textbf{C}hecking with Visual Evidence Necessity (\mymethod). Building on the mixed effects of visual evidence on verification performance observed in Section~\ref{section:4}, we examine whether the benefits of adaptive visual evidence use persist in practical settings through \mymethod. As illustrated in Figure~\ref{fig:pipeline}, \mymethod comprises two collaborating VLMs with distinct roles that operate on evidence retrieved from the knowledge source.

\begin{figure}[t]
\centering

\begin{subfigure}{0.95\linewidth}
\fcolorbox{black!50}{gray!5}{
  \begin{minipage}{0.95\linewidth}
  \ttfamily\small
  Your task is to determine whether the provided image evidence is necessary for verifying the given claim or clarifying the accompanying text evidence. Follow these steps: \\
  1. Analyze the claim and the text evidence to understand the context. \\
  2. Assess whether the image provides important information that is not already conveyed by the text. \\
  3. Decide whether the image is necessary for verification and justify your reasoning. \\
  Respond only with your analysis. \\
  \\
  Claim: \{claim\} \\
  Image Evidence: \{image evidence\} \\
  Text Evidence: \{text evidence\}
  \end{minipage}
}
\caption{Analyzer}
\label{fig:analyzer-prompt}
\end{subfigure}

\vspace{0.5em}

\begin{subfigure}{0.95\linewidth}
\fcolorbox{black!50}{gray!5}{
  \begin{minipage}{0.95\linewidth}
  \ttfamily\small
  Given a claim, your task is to determine the correct verdict based on the provided image evidence and text evidence. Provide a justification for your answer, then choose one of the following verdicts: `Supported', `Refuted', or `NEI' (Not Enough Information).\\
  Claim: \{claim\}\\
  Image Evidence: \{image evidence\}\\
  Image Analysis: \{analysis generated by Analyzer\}\\
  Text Evidence: \{text evidence\}
  \end{minipage}
}
\caption{Verifier}
\label{fig:multimodal-prompt}
\end{subfigure}

\caption{Prompts used in \mymethod.}
\label{fig:dynafc-prompt}
\end{figure}

\begin{itemize}
    \item \emph{Analyzer}: Assesses whether visual evidence is necessary for verifying a claim. Given a claim and the retrieved textual and visual evidence, it generates a natural-language assessment of visual evidence necessity together with its reasoning traces. Building on the findings in Section~\ref{section:4}, the Analyzer is instructed to reason about whether visual evidence is required for verification.
    \item \emph{Verifier}: Predicts the veracity of the claim. In addition to the claim and retrieved evidence, the Verifier incorporates the Analyzer's natural-language assessment to adaptively decide how visual evidence should be used during verification.
\end{itemize}

This modular design is inspired by the fact-checking practices of human experts~\cite{borel2023chicago,graves2019fact}, who assess the relevance and quality of candidate evidence before reaching a verdict. We hypothesize that incorporating the Analyzer's natural-language assessment can improve overall fact-checking accuracy. This hypothesis is evaluated empirically in Section~\ref{section:6}. 

Focusing on open-weight models, our best-performing configuration employs Llama-3.2-V as the Analyzer and Qwen2-VL as the Verifier. The prompts used in this framework are shown in Figure~\ref{fig:dynafc-prompt}. Both the Analyzer and the Verifier were fine-tuned on the MOCHEG training split via QLoRA~\cite{dettmers2023qlora}. The Analyzer was fine-tuned using outputs distilled from GPT-4o, generated by prompting GPT-4o with the corresponding claims and gold evidence. Further details are provided in Appendix~\ref{section:configurations}.

\section{Evaluation Results}
\label{section:6}

This section presents experimental results evaluating whether the adaptive use of visual evidence operationalized by \mymethod is effective for multimodal fact verification.

\subsection{Experimental Setting}

We evaluated fact verification performance using accuracy (ACC) and macro F1, which are standard metrics for multi-class classification. We adopted the retriever provided with MOCHEG and employed by the baseline methods to ensure a fair comparison. Specifically, a fine-tuned CLIP~\cite{clip} was used for image retrieval, while SBERT~\cite{sbert} was used for text retrieval. The top-ranked textual and visual evidence items based on cosine similarity were passed to subsequent verification stages. Additional experimental details are provided in Appendix~\ref{section:configurations}.

\subsection{Comparison with Baseline Methods}
\label{section:6.2}

On MOCHEG, we compared \mymethod with four existing multimodal fact-verification approaches. Our focus was on methods that (1) assume a fixed knowledge source $K$ and (2) employ open-weight models. For a fair comparison, all methods were trained on the MOCHEG training split.

\begin{itemize}
    \item MOCHEG~\cite{mocheg}: This method encodes the concatenated claim, textual evidence, and visual evidence using CLIP. It then applies stance detection to derive claim--text and claim--image stance features, which are averaged and fed into a multimodal fusion classifier to predict the final verdict.
    \item LVLM4FV~\cite{misinformation}: This method re-ranks initially retrieved evidence using Mistral-7B~\cite{Jiang2023Mistral7} for textual evidence and InstructBLIP~\cite{dai2023instructblip} for visual evidence. The re-ranked evidence is then fed into a LLaVa-based verifier~\cite{liu2023visual}.
    \item HGTMFC~\cite{Pang_Li_Zhang_Wang_Zhang_2025}: This method retrieves textual evidence using SBERT and visual evidence using CLIP. It encodes the claim and textual evidence as text tokens and the visual evidence as image patches using CLIP. A Hypergraph Transformer with group attention and line-graph propagation performs higher-order reasoning across evidence from different modalities, and an MLP classifier predicts the final verdict.
    \item MetaSumPerceiver~\cite{metasumperceiver}: This method encodes textual evidence in chunks using BART and visual evidence using CLIP. These modalities are fused through a dynamic Perceiver module to generate a claim-conditioned evidence summary. A fine-tuned Llama 2~\cite{touvron2023llama} then acts as a surrogate fact-checker, predicting the final veracity verdict by evaluating whether the summary is entailed by the claim.
\end{itemize}

Table~\ref{tab:baseline_mocheg_score} shows the fact-verification performance of \mymethod and the baseline methods on the MOCHEG test set. In the \emph{Gold} setting, we provided gold evidence without retrieval, simulating perfect retrieval, whereas the \emph{Retrieved} setting relied on retrieved evidence. We reported baseline scores from their original papers and omitted the macro F1 score for the retrieved setting of MetaSumPerceiver, as its implementation was not publicly available and the reported results could not be reproduced.

The results show that \mymethod consistently outperformed existing methods by a substantial margin across both evidence settings and evaluation metrics. Specifically, \mymethod achieved an accuracy of 0.612 and a macro F1 of 0.6 in the gold setting. When retrieved evidence was used, overall performance decreased, but the performance advantage of \mymethod persisted. These results demonstrate the effectiveness of the proposed framework and suggest that the adaptive use of visual evidence is beneficial for multimodal fact verification.

\begin{table}[t]
\centering
\resizebox{0.99\linewidth}{!}{%
\begin{tabular}{lcccc}
\toprule
\multirow{2}{*}{\textbf{Method}} 
& \multicolumn{2}{c}{\textbf{Gold}} 
& \multicolumn{2}{c}{\textbf{Retrieved}} \\
\cmidrule(lr){2-3} \cmidrule(lr){4-5}
& ACC & F1 & ACC & F1 \\
\midrule
\mymethod & \textbf{0.612} & \textbf{0.600} & \textbf{0.546} & \textbf{0.540} \\\midrule 
MOCHEG & 0.520 & 0.500 & 0.456 & 0.438 \\
LVLM4FV & 0.534 & 0.535 & 0.451 & 0.450 \\ 
HGTMFC & 0.541 & 0.520 & 0.486 & 0.468 \\
MetaSumPerceiver & 0.556 & 0.482 & 0.486 & - \\
\bottomrule
\end{tabular}
}
\caption{Verification performance of \mymethod and baseline methods on the MOCHEG test set, indicating the effectiveness of the proposed framework.}
\label{tab:baseline_mocheg_score}
\end{table}

\subsection{Ablation Analysis} 
\label{section:6.3}

\noindent\textbf{Analyzer-Verifier Integration}
We evaluated whether incorporating the Analyzer's natural-language assessment into the Verifier's prompt is critical to the performance gains of \mymethod. To this end, we implemented four alternative strategies that incorporate visual evidence necessity into their designs. The first, \emph{Label-only}, treats the Analyzer as a classifier that generates a visual evidence necessity label without an accompanying explanation. Only the predicted label is incorporated into the Verifier's prompt. The second, \emph{Pre-filtering (Analyzer)}, excludes images predicted as \emph{unnecessary} by the Analyzer from the Verifier's input. The third, \emph{Pre-filtering (CLIP)}, replaces the fine-tuned Analyzer with CLIP and considers visual evidence necessary when the text--image cosine similarity exceeds a predefined threshold of 0.42, optimized on the validation set. The fourth, \emph{w/o Analyzer}, removes the Analyzer entirely.

Table~\ref{tab:baseline_ablation_ftl} reports the verification performance of these strategies, all of which employ the fine-tuned Qwen2-VL as the Verifier. The results indicate that all alternative integration strategies performed worse than \mymethod, supporting the conclusion that the Analyzer's natural-language assessment contributes substantially to its performance gains. In particular, given that both Label-only and Pre-filtering (Analyzer) retain a separate call to the Analyzer, the results suggest that relying on the Analyzer's assessment of visual evidence necessity alone does not explain the observed performance gain of \mymethod. The results further suggest that the Verifier does not treat the Analyzer's assessment as a binding decision for reasoning during verification.

\begin{table}[t]
\centering
\resizebox{0.99\linewidth}{!}{%
\begin{tabular}{lcccc}
\toprule
\multirow{2}{*}{\textbf{Method}} 
& \multicolumn{2}{c}{\textbf{Gold}} 
& \multicolumn{2}{c}{\textbf{Retrieved}} \\
\cmidrule(lr){2-3} \cmidrule(lr){4-5}
& ACC & F1 
& ACC & F1 \\
\midrule
\mymethod
& \textbf{0.612} & \textbf{0.600}
& \textbf{0.546} & \textbf{0.540} \\\midrule 
Label-only
& 0.599 & 0.581
& 0.517 & 0.492 \\
Pre-filtering (Analyzer)
& 0.574 & 0.551
& 0.492 & 0.460 \\
Pre-filtering (CLIP)
& 0.569 & 0.546
& 0.495 & 0.463 \\
w/o Analyzer
& 0.563 & 0.537
& 0.477 & 0.435 \\
\bottomrule
\end{tabular}
}
\caption{Ablation results for different Analyzer--Verifier integration strategies, demonstrating the effectiveness of incorporating the Analyzer's natural-language assessment used in \mymethod.}
\label{tab:baseline_ablation_ftl}
\end{table}

\begin{table}[t]
\centering
\resizebox{0.99\linewidth}{!}{%
\begin{tabular}{lcccc}
\toprule
\multirow{2}{*}{\textbf{Method}}
& \multicolumn{2}{c}{\textbf{Gold}}
& \multicolumn{2}{c}{\textbf{Retrieved}} \\
\cmidrule(lr){2-3} \cmidrule(lr){4-5}
& ACC & F1
& ACC & F1 \\
\midrule
\mymethod
& \textbf{0.612} & \textbf{0.600}
& \textbf{0.546} & \textbf{0.540} \\\midrule 
Unified Verifier
& 0.577 & 0.555
& 0.531 & 0.513 \\
Verifier with CoT
& 0.553 & 0.511
& 0.493 & 0.459 \\
\bottomrule
\end{tabular}
}
\caption{Comparison with alternative verifiers, demonstrating the benefit of separating evidence assessment from verdict prediction.}
\label{tab:baseline_ablation_alt}
\end{table}

\noindent\textbf{Alternative Verifiers}
We compared \mymethod with two fine-tuning alternatives that use different Verifier configurations. The first, \emph{Unified Verifier}, combines the roles of the Analyzer and Verifier into a single model that jointly determines the necessity of visual evidence and predicts the veracity label. The second, \emph{Verifier with CoT}, generates a justification before predicting the verdict, unlike the Verifier in \mymethod, which predicts only the verdict. Both models are based on Qwen2-VL and were fine-tuned on the MOCHEG training split using GPT-4o-generated explanations; \emph{Unified Verifier} used the same supervision as the Analyzer in \mymethod. Table~\ref{tab:baseline_ablation_alt} reports the experimental results. \emph{Unified Verifier} underperformed relative to \mymethod, highlighting the benefit of separating evidence assessment from verdict prediction. \emph{Verifier with CoT} also underperformed, indicating that the Analyzer's assessment of visual evidence necessity provides a distinct signal beyond the benefits of eliciting explicit reasoning within the Verifier itself.

Details of the methods are described in Appendix~\ref{section:ablation_details}, and additional ablation results on the choice of VLM backbones and retriever models are presented in Appendix~\ref{section:supplementary_results}.

\subsection{Error Analysis}
\label{section:6.4}

\begin{figure}[h]
    \centering
    \includegraphics[width=.98\linewidth]{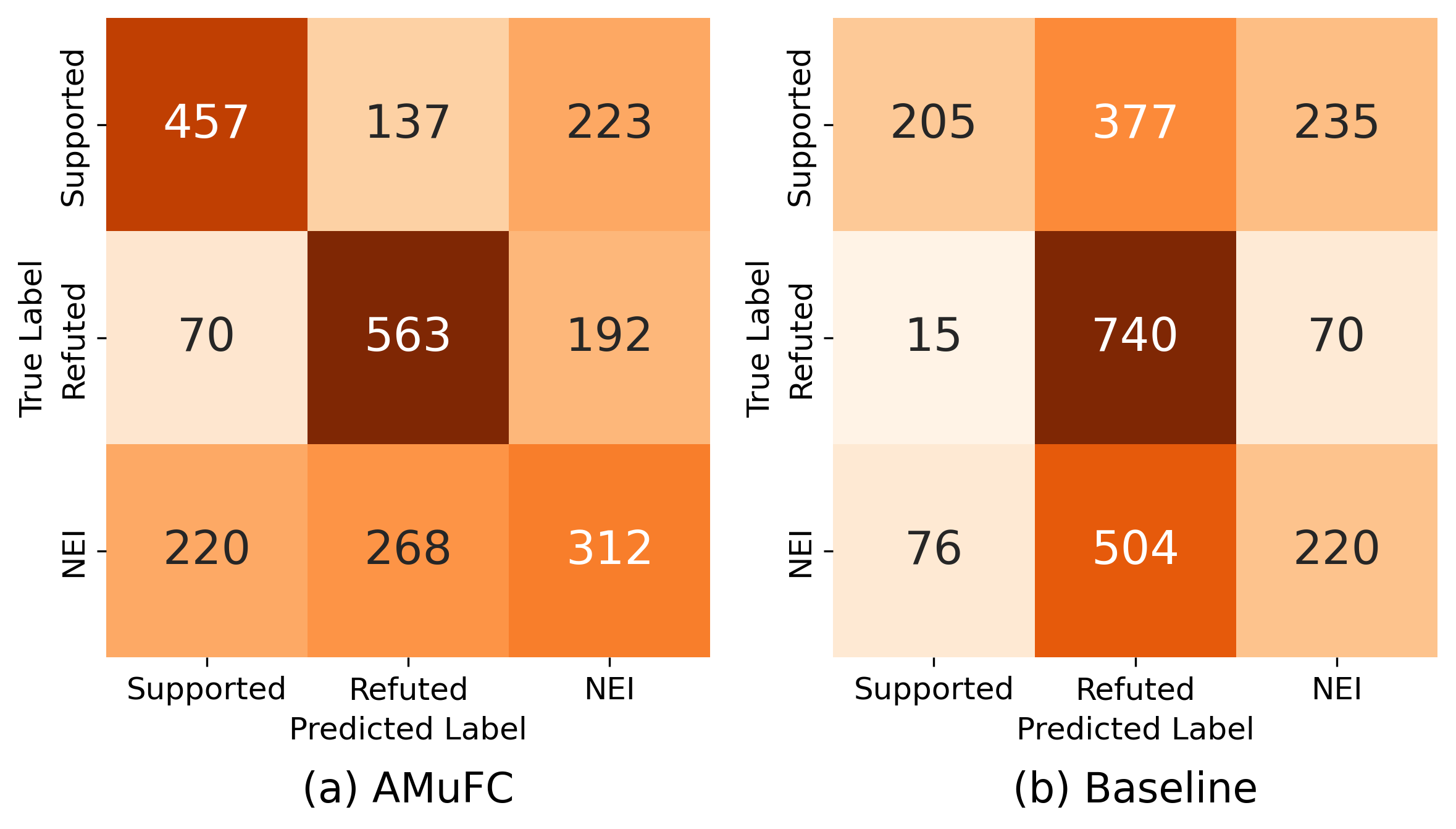}
\caption{Distribution of confusion patterns for \mymethod compared with the Verifier-only baseline.}
    \label{fig:confusion-patterns-heatmap}
\end{figure}

To assess the strengths and weaknesses of \mymethod, we compared its error distribution with that of a baseline method that excludes the Analyzer. Figure~\ref{fig:confusion-patterns-heatmap} shows the corresponding confusion matrices. Overall, the baseline method tended to overpredict the \emph{Refuted} label across the three ground-truth classes, whereas \mymethod produced more correct predictions across all classes. The largest gap was observed for the \emph{Supported} class (457 vs.\ 205). As illustrated by the left example in Table~\ref{tab:baseline-error-analysis}, the Analyzer's natural-language assessment of visual evidence necessity helped the Verifier produce the correct \emph{Supported} verdict. Another notable trend is that \mymethod predicted \emph{NEI} more frequently than the baseline, resulting in more correct \emph{NEI} predictions (312 vs.\ 220). Although this conservative tendency can sometimes lead to incorrect predictions (e.g., the right example in Table~\ref{tab:baseline-error-analysis}), such behavior may be desirable in practical fact-checking settings, where withholding a verdict in the absence of sufficient evidence can help avoid premature conclusions.

\subsection{Test-only Evaluation}
\label{section:6.5}

We conducted test-only evaluations on FIN-FACT and WebFC to assess the generalizability of the proposed framework across diverse fact-checking scenarios. We evaluated how \mymethod, trained on the MOCHEG training split, performed on these two datasets. Thus, these datasets represent practical fact-checking scenarios involving cross-domain verification and verification of recent claims using noisy web evidence, respectively. Using the same retriever for each method, we compared \mymethod with a Verifier-only baseline. Table~\ref{tab:case_study} shows that \mymethod outperformed the baseline on both datasets. While the absolute performance decreased compared with performance on MOCHEG, the performance advantage of the proposed method over the baseline persisted in both test-only evaluations without in-domain training. These results suggest that the adaptive use of visual evidence, as operationalized by \mymethod, can be effective across diverse fact-checking scenarios, including cross-domain verification and verification of recent claims using noisy web evidence.

\section{Conclusion}

This study examines a common assumption in multimodal fact-checking~\cite{misinformation,mmfc-vlm}: that incorporating visual evidence leads to more accurate verdict prediction. 
Performance analyses across different modality settings show that indiscriminate use of visual evidence can reduce fact-checking accuracy, while our manual annotation further shows that visual evidence can be particularly harmful when it is redundant or irrelevant. This observation aligns with findings from psychology research showing that some images are ineffective for misinformation correction~\cite{whitehead2025pictures}. 

Building on this finding, we present \mymethod, a modular framework that adaptively uses visual evidence through collaboration between two vision-language models with distinct roles. We conduct experiments on three datasets, including WebFC, introduced in this study. The results demonstrate the effectiveness of adaptive visual evidence use across diverse fact-checking scenarios, including cross-domain verification and verification of recent claims using noisy web evidence. Future work could extend \mymethod to agentic fact-checking systems, building on recent work demonstrating the potential of this direction~\cite{braun2025defame,lin-etal-2025-fact,jung-etal-2026-villain}.

\begin{table}[t]
\centering
\resizebox{0.99\linewidth}{!}{%
\begin{tabular}{lcccc}
\toprule
\multirow{2}{*}{\textbf{Method}} 
& \multicolumn{2}{c}{\textbf{FIN-FACT}} 
& \multicolumn{2}{c}{\textbf{WebFC}} \\
\cmidrule(lr){2-3} \cmidrule(lr){4-5}
& ACC & F1 
& ACC & F1 \\
\midrule
\mymethod
& \textbf{0.491} & \textbf{0.461}
& \textbf{0.316} & \textbf{0.241} \\

Verifier-only
& 0.469 & 0.384 
& 0.309 & 0.179 \\
\bottomrule
\end{tabular}
}
\caption{Test-only evaluation on two datasets, demonstrating the effectiveness of \mymethod across diverse fact-checking scenarios.}
\label{tab:case_study}
\end{table}

\section*{Limitations}
This work has several limitations, which open directions for future research.   First, to ensure a fair comparison with existing methods, we used the same retrievers, CLIP and SBERT, for the main experiments on MOCHEG. To address the limitation, we conducted additional experiments with different retrievers, the results of which are reported in Table~\ref{tab:baseline_ablation_ftl_concat_s}. Second, the Analyzer-Verifier collaboration introduces additional computational costs compared to the single-model baseline (Table~\ref{tab:app:comparison_tradeoff}). Employing smaller models represents a promising direction for improving the practicality of real-world fact-checking systems~\cite{belcak2025small}. 
Third, the primary evaluation was conducted on a single English-language dataset, MOCHEG. We conducted test-only evaluations using FIN-FACT and the proposed WebFC, further demonstrating the effectiveness of the proposed framework across diverse fact-checking scenarios. Future studies could incorporate recently released benchmarks that meet our data selection criteria, including VeriTaS~\cite{rothermel-etal-2026-veritas}. 

\section*{Ethical Considerations}
\paragraph{Annotator Information} Two annotators---one graduate and one undergraduate student (both male) from the authors' institution---were recruited for manual data annotation (Section~\ref{section:4}). As fact verification is an objective task, we expect that the annotators' gender had minimal impact on the outcomes. In compliance with local wage regulations, the annotators were compensated at a rate of approximately USD 7 per hour. 

\paragraph{Copyright and Privacy Issues} WebFC is released to support the evaluation of multimodal fact-checking methods. Its intended use is to evaluate methods without additional training, including zero-shot evaluation of LLM/VLMs. The dataset is released exclusively for academic purposes through \url{https://github.com/ssu-humane/AMuFC} under CC BY-NC 4.0. This dataset contains references to fact-checking articles published by PolitiFact, a service of The Poynter Institute. The CC BY-NC 4.0 license applied to WebFC does not apply to any PolitiFact content. Since the dataset is derived from publicly available fact-checking verdicts, its use poses minimal privacy concerns.

\paragraph{AI Assistant Usage} We used AI-assisted language editing tools (e.g., Grammarly and ChatGPT) exclusively to improve grammar and readability.

\section*{Acknowledgments}

This research was supported by the IITP (Institute of Information \& Communications Technology Planning \& Evaluation), funded by the Korean government (MSIT) (IITP-2026-RS-2022-00156360, IITP-2026-RS-2024-00430997, IITP-2026-RS-2020-II201602). Kunwoo Park is the corresponding author.

\bibliography{acl_latex}

\appendix

\setcounter{table}{0}
\setcounter{figure}{0}
\renewcommand\thefigure{A\arabic{figure}} 
\renewcommand\thetable{A\arabic{table}} 

\section*{Appendix}
\label{section:appendix}

\section{Experimental Details}
\label{section:configurations}

\subsection{Training and Inference}
\label{section:training_inference}
For experiments on MOCHEG, we used the fine-tuned CLIP and SBERT retrievers provided with the baseline to ensure a fair comparison by controlling for the effects of the retrieval components. We accessed GPT-4o and Gemini-2.5-Pro via API, while Qwen2-VL-7B and Llama-3.2-11B-Vision were used through pretrained checkpoints. Fine-tuning was performed with QLoRA~\cite{dettmers2023qlora}, using a rank of 128, an alpha value of 256, and a dropout rate of 0.05. We employed the AdamW optimizer with a cosine learning rate scheduler, a base learning rate of $2\times 10^{-5}$, a warmup ratio of 0.03, and a batch size of 32. Training was conducted for two epochs. All experiments were run on a single node equipped with eight NVIDIA H100 80GB GPUs, which were used for both training and inference. Training the Analyzer (Llama-3.2-11B-Vision) and Verifier (Qwen2-VL-7B) required approximately 6 and 2 H100 GPU-hours, respectively. Per-sample inference times are reported in Table~\ref{tab:app:comparison_tradeoff}.

The Verifier was fine-tuned on the MOCHEG training split using the claim, gold textual evidence, and gold image evidence when available. The training target consisted solely of the verdict label; Analyzer outputs were not used during Verifier training. The Analyzer's natural-language assessment was incorporated only at inference time through the \emph{Image Analysis} field. The same fine-tuned Verifier was used for both gold-evidence and retrieved-evidence evaluations. This decoupling also allowed different Analyzers to be paired with the same Verifier without retraining, as shown in Table~\ref{tab:baseline_ablation_model}.

We fixed the inference hyperparameters to ensure reproducibility. Specifically, we used greedy decoding for both Qwen2-VL-7B and Llama-3.2-11B-Vision; accordingly, all results from the open-weight models are deterministic. GPT-4o and Gemini-2.5-Pro were evaluated with a temperature of 0.0, with a thinking budget of 128 for Gemini-2.5-Pro. For API-based results, we reported the mean and standard error of performance across three runs. Variability due to fine-tuning seeds is evaluated separately in Appendix~\ref{section:seed_variance}.

The model IDs of the VLMs used in our experiments are provided below.

\begin{itemize}
    \item \textbf{GPT-4o~\cite{hurst2024gpt}}: gpt-4o-2024-11-20
    \item \textbf{Gemini-2.5-Pro~\cite{comanici2025gemini}}: gemini-2.5-pro
    \item \textbf{Qwen2-VL-7B~\cite{wang2024qwen2}}: \url{https://huggingface.co/Qwen/Qwen2-VL-7B-Instruct}
    \item \textbf{Llama-3.2-11B-Vision~\cite{grattafiori2024llama}}: \url{https://huggingface.co/meta-llama/Llama-3.2-11B-Vision-Instruct} 
\end{itemize}

The model IDs of the retrievers used in our experiments are provided below.

\begin{itemize}
    \item \textbf{CLIP~\cite{clip}}: \url{https://huggingface.co/sentence-transformers/clip-ViT-B-32}
    \item \textbf{SigLIP~\cite{tschannen2025siglip}}: \url{https://huggingface.co/google/siglip-so400m-patch14-384}
    \item \textbf{SBERT~\cite{sbert}}: \url{https://huggingface.co/sentence-transformers/multi-qa-MiniLM-L6-cos-v1}
    \item \textbf{mxbai~\cite{emb2024mxbai}}: \url{https://huggingface.co/mixedbread-ai/mxbai-embed-large-v1}
\end{itemize}

\subsection{Methods in Ablation Studies}
\label{section:ablation_details}

\begin{figure}[t]
\centering
\fcolorbox{black!50}{gray!5}{
  \begin{minipage}{0.95\linewidth}
  \ttfamily\small
  Your task is to determine if the provided image evidence is essential to verify the given claim or clarify the provided text evidence. To do this, follow these steps: \\
  1. Analyze the claim and the text evidence to understand the context. \\
  2. Assess whether the image evidence provides critical information not conveyed by the text alone. \\
  3. Decide if the image evidence is necessary for verification or clarification. \\
  Respond only with `Yes' if the image evidence is necessary or `No' if it is not.
  \end{minipage}
}
\caption{Prompt used for Pre-filtering (Analyzer).}
\label{fig:selector-prompt}
\end{figure}

\renewcommand\tabularxcolumn[1]{m{#1}}
\begin{table*}[t]
\centering
\setlength{\tabcolsep}{4pt}
\renewcommand{\arraystretch}{1.2}
\renewcommand\tabularxcolumn[1]{m{#1}}
\begin{tabularx}{\textwidth}{c X X}
\toprule
\textbf{Type} & \makecell[c]{\textbf{Claim ID 28}} & \makecell[c]{\textbf{Claim ID 35}} \\
\midrule

\textbf{Claim} &
Google PhoneBook &
Barack Obama's Net Worth \\
\midrule

\textbf{Refined Claim} &
Entering a phone number into the Google search engine can produce a home address and a map with directions to that address. &
Barack Obama's net worth increased over \$10 million from 2008 to 2012.\\
\bottomrule
\end{tabularx}
\caption{Examples of refinement results for incomplete claims from FIN-FACT.}
\label{tab:claim_refinement_example}
\end{table*}

\noindent\textbf{Label-only}
A VLM Analyzer (Llama-3.2-11B-Vision) was fine-tuned to predict a binary label indicating whether visual evidence is necessary, using labels generated by GPT-4o. The predicted label was then passed to the Verifier through the \emph{Image Analysis} field. The prompt is shown in Figure~\ref{fig:selector-prompt}.

\noindent\textbf{Pre-filtering (Analyzer)}
The VLM Analyzer used for \emph{Label-only} was also used to exclude images predicted to be unnecessary from the Verifier's input.

\paragraph{Pre-filtering (CLIP)}
The fine-tuned CLIP model provided by \citet{mocheg} was used to compute the cosine similarity between the claim text and the image. Images with similarity scores below 0.42, a threshold optimized on the validation set, were excluded from the Verifier's input.

\paragraph{Unified Verifier}
A VLM (Qwen2-VL-7B) was fine-tuned to perform both visual evidence necessity assessment and verdict prediction in a single pass, using a concatenated Analyzer and Verifier prompt. The model was trained to generate a natural-language analysis followed by the verdict. The analysis was distilled from GPT-4o and was identical to the supervision used to train the Analyzer in \mymethod.

\paragraph{Verifier with CoT}
A single Qwen2-VL-7B Verifier used the same prompt as the Verifier in \mymethod but was trained to generate a justification before the verdict, rather than outputting only the verdict label. The justification data were obtained from the GPT-4o-generated justifications provided by \citet{lee-etal-2024-train}.

\subsection{Licenses and Software}
\label{section:licenses_software}

\paragraph{Artifact Licenses} All datasets and open-weight models used in this study are publicly available under licenses permitting academic research use. MOCHEG~\cite{mocheg} is released under CC BY 4.0, and FIN-FACT~\cite{rangapur2025fin} is released under the MIT License. Qwen2-VL-7B is distributed under the Apache 2.0 License, whereas Llama-3.2-11B-Vision is governed by the Llama 3.2 Community License. CLIP is released under the MIT License; SBERT, SigLIP, and mxbai-embed-large are distributed under the Apache 2.0 License. The proposed WebFC dataset will be publicly released through a software repository under the CC BY-NC 4.0 for the authors' original annotations, labels, and metadata. To avoid redistributing third-party content, the release will exclude raw web documents, images, and generated document summaries, and will instead provide retrieval queries, evidence URLs, metadata, and scripts to recollect evidence from the original sources.

\paragraph{Software and Packages} All experiments were conducted using Python 3.11.13 with CUDA 12.6, using PyTorch 2.7.0, HuggingFace Transformers 4.57.3, PEFT 0.17.1, TRL 0.16.1, BitsAndBytes 0.47.0, Accelerate 1.6.0, DeepSpeed 0.17.5, SentenceTransformers 5.1.2, FAISS 1.12.0, \texttt{bm25s} 0.2.14, scikit-learn 1.7.1, and SciPy 1.16.1. Classification metrics, including accuracy and macro F1, were computed with \texttt{sklearn.metrics}, where macro F1 used \texttt{average="macro"}. Statistical tests used \texttt{scipy.stats.mannwhitneyu} with a two-sided alternative and \texttt{scipy.stats.chi2\_contingency}. Krippendorff's $\alpha$ was computed manually using the nominal-metric formulation on the 112 doubly annotated samples.

\section{Dataset Details}
\label{section:dataset_detail}

This section describes the dataset sanitization and construction processes.

\subsection{FIN-FACT}

The original FIN-FACT~\cite{rangapur2025fin} dataset contains 3,369 claims, comprising 840 text-only claims and 2,529 claims with multimodal evidence. Among 4,285 referenced image URLs, we successfully downloaded 3,027 images. We then retained 1,741 multimodal claims for which more than one referenced image was successfully retrieved. Combined with the text-only portion, the resulting dataset contains 2,581 claims.

We observed that a substantial portion of FIN-FACT claims are incomplete phrases or keywords rather than self-contained, verifiable statements, such as ``\textit{Google PhoneBook}'' or ``\textit{Barack Obama's Net Worth}.'' To address this issue, we refined vague and incomplete claims using Gemini-2.5-Pro by providing the original claim along with its evidence, justification, and label. Examples of refined claims are presented in Table~\ref{tab:claim_refinement_example}, and the claim-refinement prompt is shown in Figure~\ref{fig:finfact-claim-refinement}.

Because FIN-FACT does not provide an external knowledge source, we constructed a knowledge source $K$ consisting of 39,850 sentences extracted from all textual evidence passages in the original dataset, together with the 3,027 downloaded images. We also transformed labels from \textit{True} and \textit{False} to \textit{Supported} and \textit{Refuted}, respectively, to align with the MOCHEG label space. We used the refined FIN-FACT dataset for retrieval-based evaluation in the experiments.

\begin{figure}[t]
\centering
\fcolorbox{black!50}{gray!5}{
  \begin{minipage}{0.95\linewidth}
  \ttfamily\small
  You are a fact-checking assistant. Your task is to determine whether the given claim
  is complete in intention and clearly stated. If the claim is vague or incomplete
  (e.g., a keyword like ``Google PhoneBook''), refine it into a clear and complete
  sentence using the provided evidence, justification, and label. If the claim is
  already clear and complete, return ``not needed''.\\[0.5em]

  Claim: \{claim\}\\
  Evidence: \{evidence\}\\
  Justification: \{justification\}\\
  Label: \{label\}\\[0.5em]

  Return \textbf{ONLY} the refined claim or ``not needed''.
  \end{minipage}
}
\caption{Prompt used for claim refinement.}
\label{fig:finfact-claim-refinement}
\end{figure}

\subsection{WebFC}

Using a web-based retriever (Google Custom Search API), we queried the web for each claim to retrieve the top 10 documents and the top 1 image as textual and visual evidence. We excluded claims for which more than eight of the top 10 document URLs could not be parsed. As the retrieved documents varied widely in length and were often long, we summarized each of the documents using GPT-4o-mini (gpt-4o-mini-2024-07-18) with the prompt in Figure~\ref{fig:webfc-doc-summarization}, which is adapted from \citet{hero2}. After temporal and source filtering, 2,365 evidence remained for 621 claims.

\begin{figure}[t]
\centering
\fcolorbox{black!50}{gray!5}{
  \begin{minipage}{0.95\linewidth}
  \ttfamily\small
  Your task is to read the following document carefully and summarize it into a single, coherent paragraph. Focus on capturing the main ideas and essential details without adding new information or personal opinions. \\
  Document: \\
  \{document\}
  \end{minipage}
}
\caption{Prompt used for document summarization.}
\label{fig:webfc-doc-summarization}
\end{figure}

\section{Supplementary Results}
\label{section:supplementary_results}

\subsection{VLM Comparison}
Table~\ref{tab:baseline_ablation_model} reports verification accuracies under different VLM configurations. For the Verifier, we compared two open-weight VLMs: Qwen2-VL and Llama-3.2-V. For the Analyzer, we additionally evaluated two closed-weight VLMs accessed via API: GPT-4o and Gemini-2.5-Pro. Although our primary focus is on open-weight VLMs, this comparison allows us to assess how effectively the fine-tuned Analyzer in \mymethod performs relative to high-performing closed-weight models.

\begin{table}[t]
\centering
\resizebox{0.98\linewidth}{!}{%
\begin{tabular}{llcccc}
\toprule
\multirow{2}{*}{\textbf{Verifier}} & \multirow{2}{*}{\textbf{Analyzer}} 
& \multicolumn{2}{c}{\textbf{Gold}} 
& \multicolumn{2}{c}{\textbf{Retrieved}} \\
\cmidrule(lr){3-4} \cmidrule(lr){5-6}
& & ACC & F1 & ACC & F1 \\
\midrule

\multirow{5}{*}{Qwen2-VL} 
 & None             
 & 0.563 & 0.537 & 0.477 & 0.435 \\
 & Qwen2-VL      
 & 0.610 & 0.598 & \textbf{0.547} & \textbf{0.543} \\
 & Llama-3.2-V      
 & 0.612 & 0.600 & 0.546 & 0.540 \\
 & GPT-4o           
 & \textbf{0.631} & \textbf{0.620} & 0.538 & 0.537 \\
 & Gemini-2.5-Pro   
 & 0.612 & 0.606 & 0.478 & 0.470 \\
\midrule

\multirow{5}{*}{Llama-3.2-V} 
 & None             
 & 0.491 & 0.398 & 0.423 & 0.314 \\
 & Qwen2-VL         
 & 0.555 & 0.455 & 0.513 & 0.408 \\
 & Llama-3.2-V 
 & 0.562 & 0.461 & 0.514 & 0.409 \\
 & GPT-4o           
 & 0.575 & 0.473 & 0.515 & 0.412 \\
 & Gemini-2.5-Pro   
 & 0.525 & 0.445 & 0.415 & 0.329 \\
\bottomrule
\end{tabular}
}
\caption{Comparison of verification performance across different VLM choices for the Analyzer and Verifier.}
\label{tab:baseline_ablation_model}
\end{table}

\subsection{Effects of Varying Retrievers}

We conducted an ablation study to assess the effectiveness of \mymethod across varying retrievers. For text retrievers, we considered the pretrained SBERT, the fine-tuned SBERT, and mxbai. For image retrievers, we compared pretrained CLIP, fined-tuned CLIP, and SigLIP. As shown in Table~\ref{tab:baseline_ablation_ftl_concat_s}, the effectiveness of \mymethod persisted across different retriever configurations.

\begin{table}[t]
\centering
\resizebox{0.99\linewidth}{!}{%
\begin{tabular}{cccccc}
\toprule
\multicolumn{2}{c}{\textbf{Method}}
& \multicolumn{2}{c}{\textbf{Verifier-only}} 
& \multicolumn{2}{c}{\textbf{\mymethod}} \\
\cmidrule(lr){1-2}
\cmidrule(lr){3-4} 
\cmidrule(lr){5-6}
Text & Image
& Acc. & F1 
& Acc. & F1 \\
\midrule
SBERT & CLIP
& 0.475 & 0.434
& \textbf{0.530} & \textbf{0.527} \\
Mxbai & SigLIP
& 0.502 & 0.476
& \textbf{0.521} & \textbf{0.523} \\
SBERT(FT) & CLIP(FT)
& 0.477 & 0.435
& \textbf{0.546} & \textbf{0.540} \\
\bottomrule
\end{tabular}
}
\caption{Verification performance across different text and image retriever configurations. FT denotes fine-tuned models.}
\label{tab:baseline_ablation_ftl_concat_s}
\end{table}

\subsection{Accuracy–Efficiency Trade-offs}
\label{section:comparison_tradeoff}

While the proposed framework focuses on improving verification accuracy, the Analyzer-Verifier collaboration introduces additional computational overhead. Specifically, compared to a standard retrieval-based fact-verification pipeline, it incurs extra inference costs due to the Analyzer's assessment of the necessity of visual evidence. We provide a detailed analysis of these trade-offs in Table~\ref{tab:app:comparison_tradeoff}. We measured inference time on the test dataset with gold evidence using single-instance inference and a basic inference pipeline implemented with Hugging Face Transformers, evaluated on a single NVIDIA H100 GPU.

\begin{table}[t]
\centering
\resizebox{0.99\linewidth}{!}{%
\begin{tabular}{lccc}
\toprule
\textbf{Method} & \textbf{Accuracy} & \textbf{F1} & \textbf{Inference time}\\\midrule
\mymethod & \textbf{0.612} & \textbf{0.600} & 1.531 \\
Analyzer & - & - & 1.420\\
Verifier & - & - & 0.111\\\hline
\makecell[l]{Baseline} & 0.563 & 0.537 & \textbf{0.112}\\
\bottomrule
\end{tabular}
}
\caption{Accuracy-efficiency trade-offs, with inference time averaged per sample (seconds).}
\label{tab:app:comparison_tradeoff}
\end{table}

\begin{figure}[t]
\centering
\fcolorbox{black!50}{gray!5}{
  \begin{minipage}{0.95\linewidth}
  \ttfamily\small
  Given a claim and its associated textual and visual evidence, determine whether the visual evidence is \textbf{necessary} for evaluating the claim. \\[6pt]

  \textbf{Inputs:} \\[2pt]
  \textbf{Claim:} \{claim\} \\[2pt]
  \textbf{Textual Evidence:} \{textual evidence\} \\[2pt]
  \textbf{Visual Evidence:} \{visual evidence\} \\[6pt]

  \textbf{Necessary visual evidence} refers to visual content that provides novel, complementary, or clarifying information beyond what is conveyed in the textual evidence, and that meaningfully contributes to interpreting or supporting the claim. \\[6pt]

  \textbf{Unnecessary visual evidence} refers to visual content that is irrelevant, only depicts entities, is loosely related, or is redundant with the textual evidence. \\[6pt]

  \textbf{Select one:} \\[2pt]
  $\square$ Necessary \\
  $\square$ Unnecessary
  \end{minipage}
}
\caption{Annotation guidelines for determining whether visual evidence is necessary.}
\label{fig:annotation_guideline}
\end{figure}

\subsection{Fine-tuning Seed Variability}
\label{section:seed_variance}

To assess variability across fine-tuning runs, we retrained both the Analyzer and the Verifier with two additional seeds (43 and 44), alongside the original seed (42). All three runs were evaluated using the same prompt configuration as the main experiments. Table~\ref{tab:seed_variance} reports the mean and standard error across the three seeds on MOCHEG. \mymethod consistently outperformed the Verifier-only baseline, with mean improvements of approximately 6.0--11.0 percentage points across the four metrics. Two-sided paired t-tests across seeds indicated statistically significant improvements for all four metrics ($p<0.01$).

\begin{table}[t]
\centering
\resizebox{0.99\linewidth}{!}{%
\begin{tabular}{lcccc}
\toprule
\multirow{2}{*}{\textbf{Method}}
& \multicolumn{2}{c}{\textbf{Gold}}
& \multicolumn{2}{c}{\textbf{Retrieved}} \\
\cmidrule(lr){2-3} \cmidrule(lr){4-5}
& ACC & F1 & ACC & F1 \\
\midrule
\mymethod
& $0.613\pm0.005$ & $0.600\pm0.007$
& $0.541\pm0.006$ & $0.536\pm0.003$ \\
Verifier-only
& $0.553\pm0.006$ & $0.523\pm0.008$
& $0.476\pm0.002$ & $0.426\pm0.006$ \\
\bottomrule
\end{tabular}
}
\caption{MOCHEG verification performance across three fine-tuning seeds (mean $\pm$ standard error).}
\label{tab:seed_variance}
\end{table}

\subsection{More Comparisons on WebFC}
\label{section:dynamic_baseline}

We additionally compared \mymethod with a recent method, DEFAME~\cite{braun2025defame}, on WebFC. Both systems used the same WebFC-retrieved evidence to control for differences in web search and evidence retrieval. We used Llama-4-Scout as the Verifier backbone for both systems, following the original DEFAME configuration. For \mymethod, we replaced only the Verifier backbone with Llama-4-Scout while keeping the Analyzer unchanged. Table~\ref{tab:dynamic_baseline} shows that \mymethod achieved higher accuracy and macro F1 than DEFAME under this setting. These results provide support for the effectiveness of adaptive visual evidence use.

\begin{table}[t]
\centering
\begin{tabular}{lcc}
\toprule
\textbf{Method} & \textbf{ACC} & \textbf{F1} \\
\midrule
\mymethod & \textbf{0.662} & \textbf{0.338} \\
DEFAME & 0.478 & 0.307 \\
\bottomrule
\end{tabular}
\caption{Comparison with DEFAME on WebFC using Llama-4-Scout as the Verifier backbone.}
\label{tab:dynamic_baseline}
\end{table}

\subsection{Standalone Evaluation of the Analyzer}
\label{section:analyzer_standalone}

We evaluated the Analyzer's visual-necessity classification performance on the 112 manually annotated validation examples described in Section~\ref{subsection:4.2}. We compared the fine-tuned Analyzer used in \mymethod with its GPT-4o teacher and a majority-class baseline that always predicts \emph{Unnecessary}, which accounts for 80 of the 112 examples. Table~\ref{tab:analyzer_standalone} shows that the fine-tuned Analyzer performed comparably to its GPT-4o teacher while substantially outperforming the majority-class baseline in macro F1. These results suggest that the distilled Analyzer captures meaningful visual-necessity judgments on these annotated examples. However, \mymethod does not use the Analyzer as a binding binary classifier. As shown in Table~\ref{tab:baseline_ablation_ftl}, the Label-only variant performs worse than the full natural-language assessment, suggesting that the verbalized reasoning provides additional benefits beyond the binary necessity label.

\subsection{Effects of Generation Length}
\label{section:performance_gain_sources}

Table~\ref{tab:performance_gain_sources} summarizes the ablation configurations, average generated token counts, and verification performance in the MOCHEG gold-evidence setting. Tokens were counted using each model's tokenizer. Unified Verifier and Verifier with CoT generated 110.3 and 96.6 tokens on average, respectively, compared to approximately 90.5 tokens generated across the Analyzer and Verifier in \mymethod, yet their accuracy was lower by 3.5 and 5.9 percentage points. Despite generating only a short necessity label, Label-only improved over w/o Analyzer by 3.6 percentage points in accuracy and 4.4 points in macro F1. The full natural-language assessment yielded further improvements, indicating that the performance gains cannot be explained by generation length alone.

\begin{table}[t]
\centering
\begin{tabular}{lcc}
\toprule
\textbf{Model} & \textbf{ACC} & \textbf{Macro F1} \\
\midrule
Fine-tuned Analyzer & 0.768 & 0.721 \\
GPT-4o & 0.777 & 0.738 \\
Majority class & 0.714 & 0.417 \\
\bottomrule
\end{tabular}
\caption{Visual-necessity classification performance on 112 manually annotated validation examples.}
\label{tab:analyzer_standalone}
\end{table}

\subsection{Qualitative Examples}

Table~\ref{tab:baseline-error-analysis} shows successful and failed predictions of the proposed method, along with the corresponding Analyzer outputs. The successful case shows that the Analyzer's assessment can help the Verifier correct a baseline error when the analysis explains why visual evidence complements the text evidence in verifying the claim. However, the failed case also illustrates that the Verifier can produce an incorrect verdict despite a correct analysis, indicating that accurate evidence-necessity assessment alone does not guarantee correct verification.

\begin{table*}[t]
\centering
\resizebox{0.99\textwidth}{!}{%
\begin{tabular}{lclcccc}
\toprule
\textbf{Method} & \makecell{\textbf{Separate}\\\textbf{Analyzer}} & \makecell{\textbf{Necessity}\\\textbf{Information}} & \makecell{\textbf{Generated Tokens}\\\textbf{(Analyzer)}} & \makecell{\textbf{Generated Tokens}\\\textbf{(Verifier)}} & \textbf{ACC} & \textbf{F1} \\
\midrule
\mymethod & Yes & Assessment & 85.9 & 4.6 & \textbf{0.612} & \textbf{0.600} \\
Label-only & Yes & Label & 12.1 & 4.6 & 0.599 & 0.581 \\
Pre-filtering (Analyzer) & Yes & Filter & 12.1 & 4.6 & 0.574 & 0.551 \\
w/o Analyzer & No & None & -- & 4.6 & 0.563 & 0.537 \\
Unified Verifier & No & Joint assessment & -- & 110.3 & 0.577 & 0.555 \\
Verifier with CoT & No & None & -- & 96.6 & 0.553 & 0.511 \\
\bottomrule
\end{tabular}
}
\caption{Average generation length and verification performance across ablation configurations.}
\label{tab:performance_gain_sources}
\end{table*}

\begin{table*}[t]
\small
\centering
\renewcommand{\arraystretch}{1.2}
\begin{tabularx}{\textwidth}{c X X}
\toprule
\textbf{Type} & \makecell[c]{\textbf{Successful}} & \makecell[c]{\textbf{Failed}} \\
\midrule
\textbf{Claim} 
& A Boeing B-17E bomber from World War II was found in the jungle with coffee still in thermoses.
& Students no longer say the Pledge of Allegiance in schools. \\
\midrule
\textbf{Visual Evidence} 
& \makebox[\linewidth]{\includegraphics[width=0.32\textwidth]{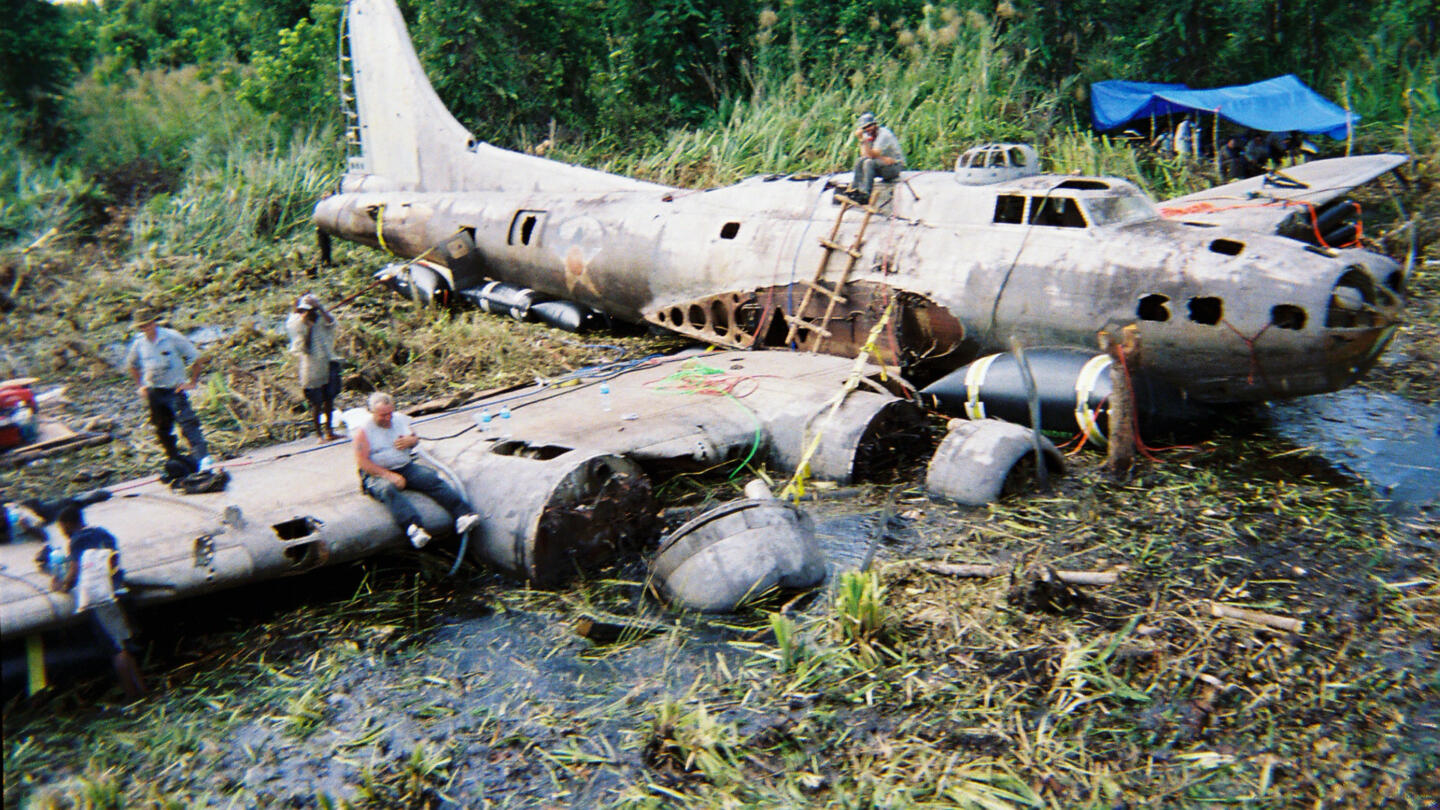}}
& \makebox[\linewidth]{\includegraphics[width=0.32\textwidth]{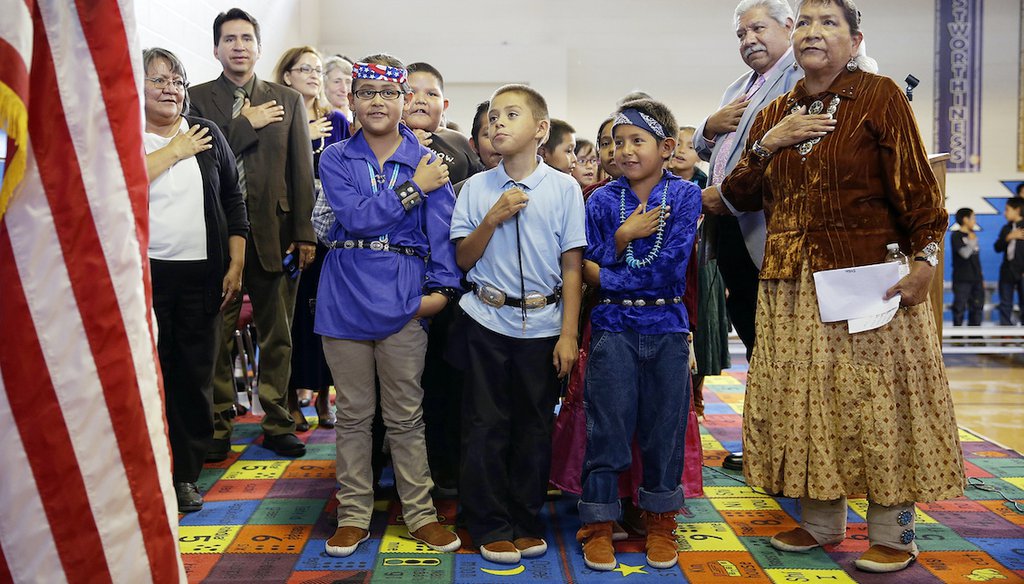}}
\\
\midrule
\textbf{Textual Evidence} 

& \hangindent=1em \noindent • They found the rear door of the airplane, two aluminum receptacles and a third which contained coffee residue.

\hangindent=1em \noindent • Both the impactor and the condensate were stored and transported to the laboratory at 2-8 °C.

\hangindent=1em \noindent • Print LONG BEACH, Calif. - A B-17 bomber that lay in a New Guinea swamp for decades after being forced down during a World War II combat mission has been returned to the United States after years of salvage efforts.

\hangindent=1em \noindent • As food stores were running low at the airplane, Zerbino and Fito went searching for cadavers and although they found one, they were too tired to bring it back to the fuselage.

\hangindent=1em \noindent • Ignited by fire, ammunition was exploding inside the cabin.

& \hangindent=1em \noindent • Students still say the Pledge of Allegiance in classrooms, though they don't have to.

\hangindent=1em \noindent • Virtual learning has presented new challenge for how schools say the Pledge of Allegiance. \\
\midrule
\textbf{Analysis} 
& The image evidence is \textbf{necessary} for verifying the claim. While the text evidence provides detailed information about the discovery of the Boeing B-17E bomber in the jungle, including the presence of coffee residue and thermoses, the image directly shows the plane in its jungle location, confirming the physical context of the claim. The visual evidence complements the text by providing a tangible representation of the scene described, which is essential for verifying the authenticity of the claim.
& The image evidence is \textbf{not necessary} for verifying the claim. The text evidence already provides sufficient information to address the claim that students no longer say the Pledge of Allegiance in schools. It clarifies that students still say the Pledge of Allegiance in classrooms, though it is not mandatory, and that it is recited at every school. The image of students raising their hands does not add any new or essential information to verify or refute the claim. \\
\midrule
\textbf{Prediction} 
& Ground truth: \textit{Supported} \newline
Baseline: \textit{Refuted} \textcolor{red}{\ding{55}} \newline
\mymethod: \textit{Supported} \textcolor{green!60!black}{\ding{51}}
& Ground truth: \textit{Refuted} \newline
Baseline: \textit{Refuted} \textcolor{green!60!black}{\ding{51}}\newline
\mymethod: \textit{NEI} \textcolor{red}{\ding{55}} \\
\bottomrule
\end{tabularx}
\caption{Examples for successful and failed predictions of \mymethod.}
\label{tab:baseline-error-analysis}
\end{table*}

\end{document}